\documentclass{article}

\usepackage{arxiv}

\usepackage[utf8]{inputenc} 
\usepackage[T1]{fontenc}    
\usepackage{hyperref}       
\usepackage{url}            
\usepackage{booktabs}       
\usepackage{amsfonts}       
\usepackage{nicefrac}       
\usepackage{microtype}      
\usepackage{lipsum}
\usepackage{graphicx}
\usepackage{amsmath}
\usepackage{float}

\graphicspath{ {./images/} }

\title{A Review on Generative AI for Text-to-Image and Image-to-Image Generation and Implications to Scientific Images}

\author{
  Zineb Sordo \\
  Applied Math and Computational Research\\
  Lawrence Berkeley National Laboratory\\
  Berkeley, CA 94720 \\
  \texttt{zsordo@lbl.gov} \\
   \And
  Eric Chagnon \\
  Applied Math and Computational Research\\
  Lawrence Berkeley National Laboratory\\
  Berkeley, CA 94720 \\
  \texttt{echagnon@lbl.gov} \\
  \And
 Daniela Ushizima \thanks{Corresponding Author}\\
  Applied Math and Computational Research\\
  Lawrence Berkeley National Laboratory\\
  Berkeley, CA 94720 \\
  \texttt{dushizima@lbl.gov} \\
}

\begin{document}
\maketitle
\begin{abstract}
This review surveys the state-of-the-art in text-to-image and image-to-image generation within the scope of generative AI. We provide a comparative analysis of three prominent architectures: Variational Autoencoders, Generative Adversarial Networks and Diffusion Models. For each, we elucidate core concepts, architectural innovations, and practical strengths and limitations, particularly for scientific image understanding. Finally, we discuss critical open challenges and potential future research directions in this rapidly evolving field.
\end{abstract}


\section{Introduction}

Generative AI (genAI) has emerged as a powerful tool with the ability to create novel digital content, including images, text, and music~\cite{foster2023generative}. However, using generative AI to create scientific images of phenomena unseen by the model continues to be challenging, and prone to hallucination~\cite{Zhou:2024} and misrepresentation of scientific principles. If the model extrapolates beyond its training data, it can generate images that, while visually plausible, are physically or biologically impossible~\cite{Sun:2024}. This can lead to the propagation of inaccurate scientific concepts and hinder genuine discovery~\cite{Lucas,Maleki:2024}. This paper overviews the major milestones in the last few years, then describes
how Variational Autoencoders (VAEs), Generative Adversarial Networks (GANs) and Diffusion Models have revolutionized these areas. Finally, we delineate potential avenues for verification and validation. 

Overall, this paper focuses on two key subdomains: text-to-image and image-to-image generation. 

This review aims to:

\begin{itemize}
\item Analyze their applications in text-to-image and image-to-image generation.
\item Compare and contrast the strengths and weaknesses of these approaches for scientific data generation.
\item Discuss the challenges and future directions of research in this field.
\end{itemize}

\section{Background}
This section provides an overview of the key advancements in text-to-image and image-to-image generation technologies, particularly considering the technologies from major tech companies, such as Google, Meta, Microsoft and OpenAI. We highlight significant software releases and the underlying algorithms that have shaped the landscape of generative AI from 2021 to 2024.

In 2021, OpenAI introduced DALL-E, a groundbreaking model that utilized a variant of the GPT-3 architecture, which is a large language model (LLM), to generate images from textual descriptions. DALL-E employed a transformer-based architecture, leveraging the principles of attention mechanisms to understand and synthesize complex relationships between text and visual elements. The model was trained on a diverse dataset of text-image pairs, enabling it to create novel images that often combined disparate concepts in coherent and imaginative ways. This marked a significant advancement in generative models, setting the stage for future developments in text-to-image synthesis\cite{ramesh2021zeroshottexttoimagegeneration,openai2021clip}.

\begin{figure}[h]
    \centering
    \includegraphics[width=0.9\linewidth]{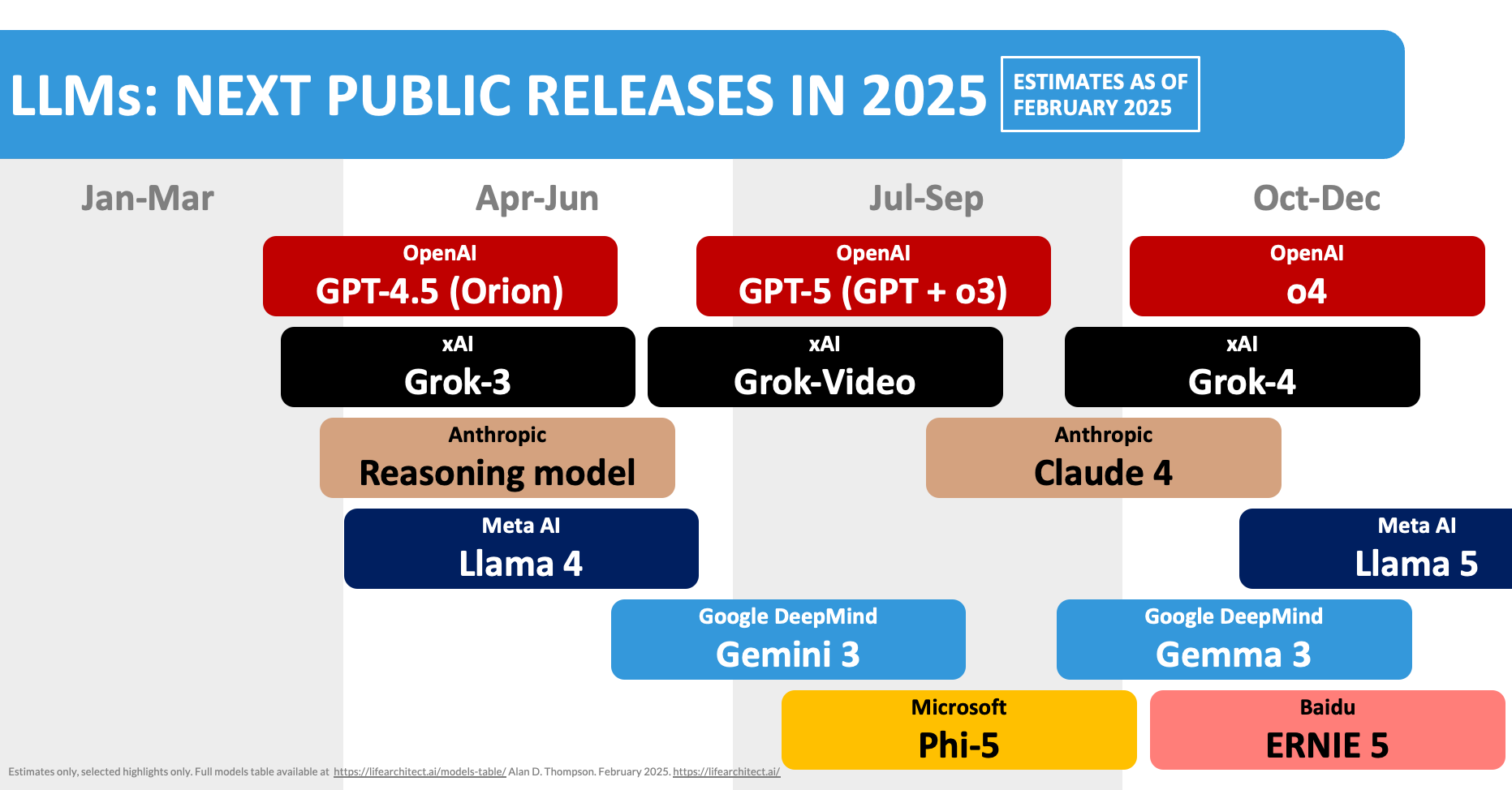}
    \caption{Major highlights of language and multimodal models, with less focus on text-to-image generation models \cite{lifearchitect}.}
    \label{fig:LLM_history}
\end{figure}

The year 2022 saw significant advancements in text-to-image generation technologies. Google introduced Imagen, a diffusion-based model that generates high-quality images from textual prompts. Imagen utilized a two-step process involving a denoising diffusion probabilistic model (DDPM), which iteratively refined a random noise image into a coherent visual representation based on the input text. While Imagen did not directly employ Variational Autoencoders (VAEs) or Generative Adversarial Networks (GANs)~\cite{Goodfellow:2016}, its diffusion approach presented an alternative to these traditional generative frameworks. Meanwhile, OpenAI released DALL-E 2, which improved upon its predecessor by incorporating CLIP (Contrastive Language-Image Pretraining)~\cite{openai2021clip} to better align the generated images with textual descriptions\cite{radford2021learningtransferablevisualmodels}. CLIP itself is a multi-modal vision and language model that understands and relates textual and visual information. CLIP has since become the benchmark for text-to-image and multi modal text+image-to-image generation. Meta's Make-A-Scene also emerged, allowing users to have more control over the composition of generated images through scene graphs, enhancing the interactivity of the image creation process\cite{Gafni_Polyak_Ashual_Sheynin_Parikh_Taigman_2022}. Additionally, Microsoft integrated OpenAI's models into its Azure platform, making these advanced capabilities more accessible to developers and businesses. 

In 2023, the landscape of generative AI continued to evolve with notable releases and innovations. Google unveiled Imagen Video, extending the capabilities of its diffusion models to generate videos from text prompts, thus introducing temporal coherence and movement to the generative process\cite{ho2022imagenvideohighdefinition}. OpenAI launched DALL-E 3, which featured enhanced text comprehension and image generation capabilities, further refining the alignment between textual input and visual output through improved training techniques and larger datasets\cite{BetkerImprovingIG}. This model continued to leverage the principles of LLMs to enhance its generative abilities. Anthropic's Claude AI also began to incorporate multimodal functionalities, allowing for a richer interaction between text and images, while not strictly focused on image generation, it illustrated the growing trend of integrating LLMs with visual understanding. Meta's Segment Anything Model (SAM), a zero-shot algorithm, has facilitated advanced image segmentation, providing tools for image-to-image manipulation by enabling users to specify regions of interest within images\cite{kirillov2023segany}. Finally, Microsoft’s Copilot integrated these generative AI technologies into design workflows, allowing users to seamlessly leverage AI-driven image generation and manipulation tools, thus democratizing access to sophisticated design capabilities\cite{Spataro_2023}.

In 2024, major tech players release improved technologies for text-to-image and image-to-image generation. Google, with its ongoing development of Imagen and related models, focused on enhanced photorealism and semantic understanding, leveraging diffusion models and potentially incorporating LLMs for improved prompt interpretation. Meta expanded its offerings, building upon its Emu architecture, emphasizing both speed and quality, exploring variations of diffusion models potentially incorporating VAEs for efficient latent space manipulation \cite{Kokhlikyan_Jayaraman_Bordes_Guo_Chaudhuri_2023}\cite{girdhar2024emuvideofactorizingtexttovideo}\cite{Sheynin_2024_CVPR}. OpenAI continued refining DALL-E, focusing on higher resolution and more consistent image generation, further optimizing its diffusion-based pipeline and potentially integrating LLM-driven refinement steps. Anthropic, while primarily known for LLMs, began exploring visual generation in conjunction with its Claude model, potentially integrating diffusion models with their sophisticated contextual understanding capabilities. Microsoft further solidified its position with updates to its Designer and Image Creator powered by DALL-E 3, focusing on user accessibility and integration within its ecosystem. Underpinning these advancements are diffusion models, which iteratively denoise images from Gaussian noise, often operating within a latent space defined by VAEs for computational efficiency. Some models, though less prevalent in 2024 for high-fidelity image generation, may still utilize GANs for specific tasks like upscaling or style transfer, though diffusion models have become the dominant approach for general text-to-image and image-to-image tasks. The integration of LLMs, especially in prompt understanding and refinement, became a key trend, ensuring generated images more accurately reflect the intent of the user.

\section{Key Generative Architectures}
\subsection{Variational Auto-Encoder (VAE)}
First introduced in 2013~\cite{kingma2022autoencodingvariationalbayes}, the Variational Auto-Encoder (VAE) is a type of generative neural network capable of learning a probability distribution over a set of data points without labels. A VAE learns to encode input data into a lower-dimensional latent space and then decode it back to the original space by sampling latents, while ensuring the latent representations follow a known probability distribution.

VAE is categorized as a model with an explicit intractable density function (tractable models allow for explicit likelihood computation) that learns a probability distribution using variational inference and latent variables. Intuitively, latent variables (LV) ``explain'' the data in a ``simpler'' way, and more rigorously, LV result from a transformation of the data points into a continuous lower-dimensional space. 
Mathematically, we define $x$ as a data point that follows a probability distribution $\mathbf{p(x)}$ and $z$, to be a latent variable that follows a probability distribution $\mathbf{p(z)}$, then:
\begin{itemize}
    \item $\mathbf{p(x)}$ is the marginal distribution (and goal of the model)
    \item $\mathbf{p(z)}$ is the prior distribution
    \item $\mathbf{p(x|z)}$ is the likelihood mapping latents $z$ to data points $x$
    \item $\mathbf{p(x,z) = p(x|z)*p(z)}$ is the joint distribution of data points and latent variables 
    \item $\mathbf{p(z|x)}$ is the posterior distribution that describes $z$ that can be produced by $x$
\end{itemize}
   
\begin{figure}[H]
    \centering
    \includegraphics[width=0.4\linewidth]{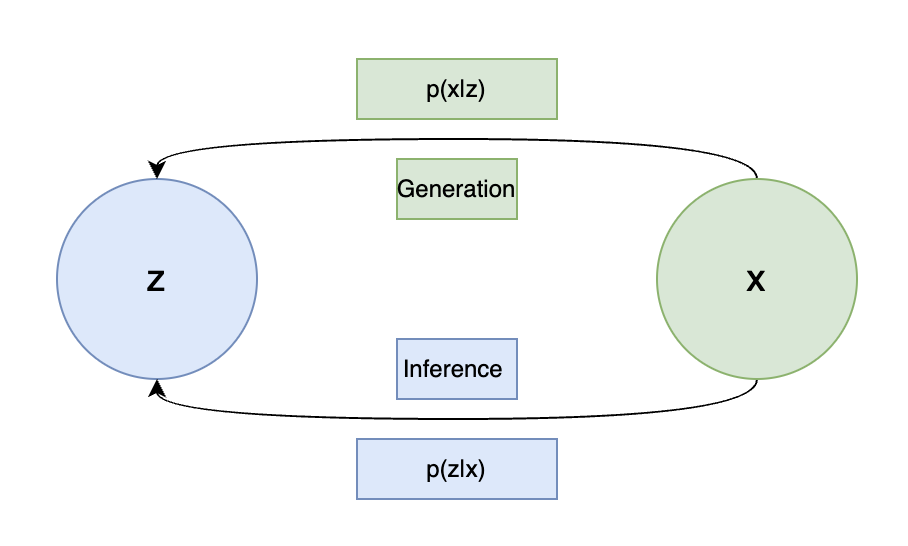}
    \caption{Variational Inference}
    \label{fig:vae_process}
\end{figure}

The generative process in VAEs consists of computing $x$ given $z$ (computed by $\mathbf{p(x|z))}$ by sampling $\mathbf{z \sim p(z)}$ and $\mathbf{x \sim p(x)}$, while inference consists of finding $\mathbf{z}$ given $\mathbf{x}$ (computed by $\mathbf{p(z|x)}$) by sampling $\mathbf{x \sim p(x)}$ and then sampling $\mathbf{x \sim p(z|x)}$ (see Figure ~\ref{fig:vae_process}).
To find the parameters of the marginal distribution $\mathbf{p(x)}$, we can apply gradient descent which translates to compute the following:
\begin{equation}
\centering
    \nabla logp_\theta(x) = \int p_\theta(z | x)\nabla_{\theta}logp_\theta(x,z)dz
\end{equation}
The goal of variational inference is to approximate the posterior distribution $\mathbf{p(z|x)}$ with an explicit tractable probability distribution and allow its computation as an optimization problem. We can call this distribution the variational posterior $\mathbf{q(z|x)}$. During training, the goal is to minimize the Kullback-Leibler (KL) divergence, which expresses the difference between the true posterior and the variational posterior and is given by (with $\theta$ being the model parameters):

\begin{equation}
    \centering
    KL(P||Q) = \int^\infty_{-\infty} p(x)log(\frac{p(x)}{q(x)})dx = E_{x \sim p(x)} [log(\frac{p(x)}{q(x, \theta)}] 
\end{equation}

To enable backpropagation in the decoder part of the VAE, this method considers a reparametrization trick where instead of just sampling from the latent distribution, they add a random noise to the mean and standard deviation to make gradient computation possible.

\begin{figure}[H]
    \centering
    \includegraphics[width=0.8\linewidth]{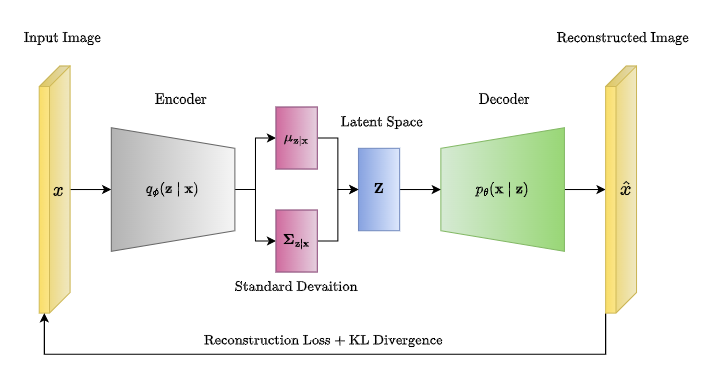}
    \caption{VAE Encode - Decoder architecture}
    \label{fig:vae-architecture}
\end{figure}

The final loss function is given by the following equation: 
\begin{equation}
    L_{\theta, \phi}(x) = E_{q_{\phi(z|x)}}[logp_\phi(x|z)] - KL(q_\phi(z|x) || p_\theta(z))
\end{equation}
The first part of the previous equation (also called negative reconstruction error associated to the Decoder part of the VAE) controls how well the model reconstructs $x$ given $z$ of the variational posterior whereas the second part (corresponding the Encoder part of the VAE) controls how close the variational posterior $\mathbf{q(z|x)}$ is to the prior $\mathbf{p(x)}$ i.e. how well the dimensionality reduction of the Encoder captures the data features within the latent space.

\subsection{Generative Adversarial Networks (GANs)}

\begin{figure}[H]
    \centering
    \includegraphics[width=0.65\linewidth]{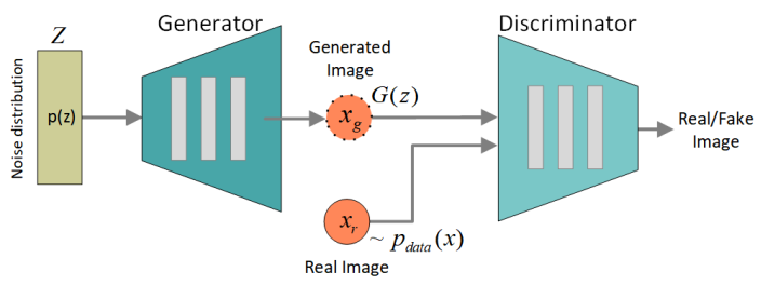}
    \caption{Architecture of the Vanilla GAN}
    \label{fig:vanillaGAN-architecture}
\end{figure}

The first Generative Adversarial Network (GAN) was introduced in 2014 ~\cite{NIPS2014_5ca3e9b1,Goodfellow:2016} and represents a major advancement in generative learning. GANs are a class of machine learning models that consists of two key components: a generator and a discriminator, where the generator aims to produce synthetic data, and the discriminator attempts to distinguish between the real data and synthetic data (see Figure ~\ref{fig:vanillaGAN-architecture}.

\begin{itemize}
\item The generator $\mathbf{G(z)}$ maps random noise $z \sim p(z)$ (also called latents) to the data distribution and outputs the synthetic image in the shape of a 1D-vector. The stochasticity given by this random sampling will provide a non-deterministic output, which is how the model creates diversity in the generation process. The goal here is to fool the discriminator and minimize $log(1-D(G(z)))$, which amounts to maximizing the discriminator's mistakes. 

\item The discriminator $\mathbf{D(x)}$ takes as input a real and synthetic image (generated by the generator) and outputs the probability that the image corresponds to the real data distribution or not. The goal here is to maximize the loss function or the probability that it correctly classifies real and fake images.
\end{itemize}

This adversarial process drives both the generator and the discriminator to improve, resulting in high-quality synthetic data. In addition, the fact that the generator is only trained to fool the discriminator makes this Vanilla GAN model unsupervised. 

The goal of the GAN is to solve the min-max game or adversarial game between the generator and the discriminator with the following objective function and optimization problem: 

\begin{equation}
         \min_{G} \max_{D} V(G, D) = \mathbb{E}_{x\sim p_{data}(x)}[logD(x)] + \mathbb{E}_{z\sim p_{fake}(z)}[log(1-D(G(z)))] 
\end{equation}

where $D(x)$ is the probability that $x$ is real, $G(z)$ is the generated sample, and thus $D(G(z))$ is the probability that the generated image given latent $z$ is real. 
One of the most common limitations of GANs is the so-called \textbf{mode collapse} problem where the generator fails to represent accurately the pixel space of all possible outputs. This issue is common in high-resolution images, where too many fine-scale features must be captured. In that case, the generator gets stuck in a parameter setting with a similar level of noise that can consistently fool the discriminator and only captures a subset of the real data distribution. It then fails to produce diversity in its outputs and collapses to producing only a few types of synthetic samples.

\subsubsection{Conditional GAN (CGAN)}
As an extension of the Vanilla GAN, the Conditional GAN was introduced in 2014 ~\cite{mirza2014conditionalgenerativeadversarialnets}  and uses conditional information (image or text) to guide the generation process. The CGAN performs conditioning generation by feeding information to both the generator and the discriminator. The generator $G(z, y)$ takes as input random noise $z$, and the conditional embedding $y$ and learns to generate data given this condition, whereas, the discriminator $D(x,y)$ learns to classify real and fake images by checking that the condition $y$ is met. The min-max optimization function becomes: 
\begin{equation}
     \min_{G} \max_{D} V(G, D) = \mathbb{E}_{x\sim p_{data}(x)}[logD(x|y)] + \mathbb{E}_{z\sim p_{fake}(z)}[log(1-D(G(z|y), y))] 
\end{equation}
StackGAN~\cite{8237891} and Attentional GAN (AttnGAN)~\cite{8578241} are influential CGAN architectures that advanced text-to-image generation. StackGAN introduced a hierarchical approach, generating low-resolution images and iteratively refining them to high-resolution outputs. AttnGAN innovated with attention mechanisms, allowing the model to selectively attend to specific words or phrases in the text description when generating corresponding image regions.

\subsubsection{Deep Convolutional GAN (DCGAN)}
Following the initial development of GANs, various architectures emerged, notably Deep Convolutional Generative Adversarial Networks (DCGANs) introduced by Radford et al. in 2015~\cite{radford2016unsupervisedrepresentationlearningdeep}, which extended the foundational GAN framework. While the Vanilla GAN's architecture contains simple downsampling and upsampling layers with ReLU activations and a Sigmoid activation for the discriminator, this variant of the GAN is made of strided convolution layers, batch norm layers, and LeakyReLU activation functions. 
This architecture is adapted to small size images such as RGB inputs of shape \textit{(3,64,64)}.

\subsection{Diffusion Models}

\begin{figure}
    \centering
    \includegraphics[width=0.9\linewidth]{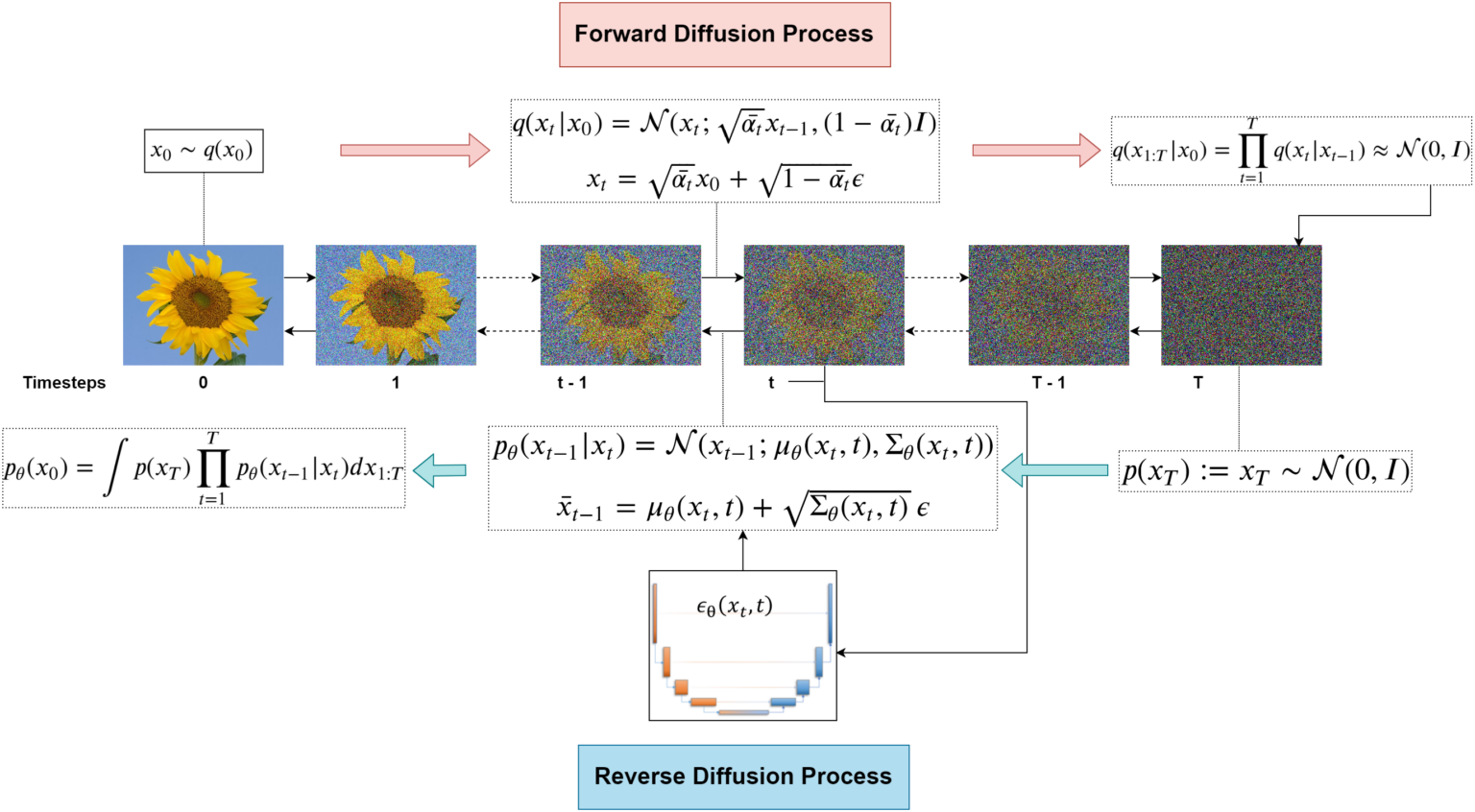}
    \caption{Denoising diffusion probabilistic models (DDPMs). Source: ~\cite{Kulkarni2023}}
    \label{fig:diffusion-model-architecture}
\end{figure}

Diffusion models, now producing state-of-the-art high-fidelity and diverse images, have evolved from the initial work of Sohl-Dickstein et al. in 2015~\cite{pmlr-v37-sohl-dickstein15} to the significantly impactful Denoising Diffusion Probabilistic Models (DDPM) by Ho et al. in 2020~\cite{NEURIPS2020_4c5bcfec}. These models differ from the previous generative models as they decompose the image generation process through small denoising steps. In fact, the idea behind diffusion models is that they take an input image $x_0$ and gradually add Gaussian noise in what is called the forward process. The second part of the network is the reverse process, or sampling process, which consists of removing the noise to obtain new data (see Figure ~\ref{fig:diffusion-model-architecture}). 

The \textit{forward} process in the diffusion network consists of a Markov chain of $T$ steps. Given an input image $x_0$ sampled from the true data distribution $x_0 \sim \mathbf{q(x_0)}$, at each step $t < T$, Gaussian noise is added to $x_{t-1}$, according to a variance schedule $\beta_1,...,\beta_t$ to obtain $x_{t} \sim \mathbf{q(x_t|x_{t-1})}$ where the $x_1, ...,x_T$ are latents of the same dimensionality as $x_0$:
\begin{equation}
    q(\mathbf{x}_t|\mathbf{x}_{t-1}) = \mathcal{N}(\mathbf{x}_t ; \mathbf{\mu}_t = \sqrt{1 - \beta_t}\mathbf{x}_{t-1}, \mathbf{\Sigma}_t = \beta_t I)
\end{equation}

Where $I$ is the identity matrix and the variances $\beta_t$ can be learned or kept constant as hyperparameters. In the case of the DDPM paper, the authors used a linear schedule increasing from $\beta_1 = 10^{-4}$ to $\beta_T = 0.02$. The scheduler, however, can be linear, quadratic, cosine ~\cite{nichol2021improved} etc. 
The posterior probability can be defined as: 
\begin{equation}
    q(\mathbf{x}_{1:T}|\mathbf{x}_{0}) = \mathbf{\prod_{t=1}^T} q(\mathbf{x}_t|\mathbf{x}_{t-1})
\end{equation}

And using the reparametrization trick to obtain a tractable closed-form sampling at any timestep, we define $\alpha_t = 1-\beta_t$, $\bar{\alpha} = \prod_{s=0}^t\alpha_s$ where $\epsilon_0,...,\epsilon_{t-1} \sim \mathcal{N}(\mathbf{0, I})$ and finally have: 
\begin{equation}
    \mathbf{x}_t \sim q(\mathbf{x}_t|\mathbf{x}_0) = \mathcal{N}(\mathbf{x}_t; \bar{\alpha}_t\mathbf{x}_0, (1-\bar{\alpha})\mathbf{I})
\end{equation}
Given that $\beta_t$ is a hyperparameter, it is possible to compute $\alpha_t$ and $\bar{\alpha}_t$ for all timesteps. We can therefore sample the noise at any timestep $t$ and get the latent variables $\mathbf{x}_t$. 

The second part of diffusion models is the \textit{reverse} process which is also a Markov chain with learned Gaussian transitions starting at $p(\mathbf{x}_T) = \mathcal{N}(\mathbf{x}_T;\mathbf{0},\mathbf{I})$. The goal of the reverse process is to learn the reverse distribution $q(\mathbf{x}_{t-1}|\mathbf{x}_t)$ by approximating it with a parametrized model $p_\theta$ (where $p_\theta$ is Gaussian and the mean and variance will be parametrized and learned by a neural network):

\begin{align}
    p_\theta (\mathbf{x}_{t-1}|\mathbf{x}_t) &= \mathcal{N}(\mathbf{x}_{t-1}; \mathbf{\mu}_\theta(\mathbf{x}_t, t), \mathbf{\Sigma}_\theta(\mathbf{\mathbf{x_t}, t)}) \\[5pt]
    p_\theta(\mathbf{x}_{0:T}) &=  p_\theta(\mathbf{x}_{T})\prod_{t=1}^T p_\theta (\mathbf{x}_{t-1}|\mathbf{x}_t) 
\end{align}

To train diffusion models, we optimize the negative log-likelihood of the training data. This optimization, similar to the approach used in Variational Autoencoders (VAEs), is achieved by maximizing the Evidence Lower Bound (ELBO), a tractable approximation. The following expression represents the ELBO after a series of computational steps:

\begin{align}
\log p(\mathbf{x}) &\geq \mathbb{E}_{q(\mathbf{x}_1 | \mathbf{x}_0)} \left[\log p_{\theta}(\mathbf{x}_0 | \mathbf{x}_1)\right] - D_{KL} \left(q(\mathbf{x}_T | \mathbf{x}_0) \| p(\mathbf{x}_T) \right) - \sum_{t=2}^{T} \mathbb{E}_{q(\mathbf{x}_t | \mathbf{x}_0)} 
\left[ D_{KL} \left(q(\mathbf{x}_{t-1} | \mathbf{x}_t, \mathbf{x}_0) \| p_{\theta}(\mathbf{x}_{t-1} | \mathbf{x}_t) \right) \right] \\ 
\log p(\mathbf{x}) &\geq  L_0 - L_T - \sum_{t=2}^{T} L_{t-1}
\end{align}
\vspace{-0.2cm}
Where:
\begin{itemize}
    \item $\mathbb{E}_{q(\mathbf{x}_1 | \mathbf{x}_0)} \left[\log p_{\theta}(\mathbf{x}_0 | \mathbf{x}_1)\right]$ is the reconstruction term
    \item $D_{KL} \left(q(\mathbf{x}_T | \mathbf{x}_0) \| p(\mathbf{x}_T) \right)$ basically shows how close $\mathbf{x}_T$ is to the Standard Gaussian distribution. 
    \item $\sum_{t=2}^{T} L_{t-1} = L_t$ represents the difference between the denoising steps $p_{\theta}(\mathbf{x}_{t-1}|\mathbf{x}_t)$ and the approximated ones $q_{\theta}(\mathbf{x}_{t-1}|\mathbf{x}_t, \mathbf{x}_0)$. The KL divergence compares $p_{\theta}(\mathbf{x}_{t-1}|\mathbf{x}_t)$ against forward process posteriors, which are tractable when conditioned on $\mathbf{x}_0$. 
\end{itemize}

Therefore maximizing the likelihood amounts to learning the denoising steps $L_t$. Based on further calculations, the DDPM paper shows that instead of predicting the mean of the distribution during the training of the reverse process, the model will predict the noise $\mathbf{\epsilon}$ at each timestep $t$ using the following simplified formula of the denoising term in the ELBO (also the loss function of the reverse process network):

\begin{equation}
    L_{\text{simple}}(\theta) := \mathbb{E}_{t, x_0, \epsilon} 
\left[ \|\epsilon - \epsilon_{\theta} (\sqrt{\bar{\alpha}_t} x_0 + \sqrt{1 - \bar{\alpha}_t} \epsilon, t) \|^2 \right]
\end{equation}

The model associated to the loss function of the reverse process is a U-Net architecture with residual blocks, group normalization as well as self-attention blocks. The timestep $t$ is concatenated to the input image using a cosine positional embedding into each residual block. This denoising U-Net is trained to predict the noise at each timestep of the process. 

\subsubsection{Conditional Image Generation with Guided Diffusion}

\textbf{Classifier Guidance}

Similarly to the CGAN, an important extension of the diffusion model is the \textit{Guided diffusion} model that includes conditional image generation in the network. In that scenario, the model adds conditioning information $y$ at each diffusion step: 
\begin{equation}
    p_\theta(\mathbf{x}_{0:T}| y) =  p_\theta(\mathbf{x}_{T})\prod_{t=1}^T p_\theta (\mathbf{x}_{t-1}|\mathbf{x}_t, y) 
\end{equation}

Using Bayes rule with some computations and more importantly by adding the guidance scalar term $s$, we can show that guided diffusion models aim to learn $\nabla logp_{\theta}(\mathbf{x}_t|y)$ such that: 
\begin{equation}
    \nabla logp_{\theta}(\mathbf{x}_t|y) = \nabla logp_{\theta}(\mathbf{x}_t) + s.\nabla logp_{\theta}(y|\mathbf{x}_t)
\end{equation}

It was also shown in ~\cite{pmlr-v37-sohl-dickstein15} and ~\cite{NEURIPS2021_49ad23d1} that a \textbf{classifier guidance} model defined by $f_\Phi(y|\mathbf{x}_t, t)$ can guide the diffusion towards the target class $y$ by training $f_\Phi(y|\mathbf{x}_t, t)$ on a noisy image $\mathbf{x}_t$ to predict class $y$. To do so, we build a class-conditional diffusion model with mean $\mu(\mathbf{x}_t|y)$ and variance $\Sigma_{\theta}(\mathbf{x}_t|y)$ and perturb the mean by the gradients of $logf_\Phi(y|\mathbf{x}_t, t)$ of class y, resulting in: 

\begin{equation}
    \hat{\mu}(\mathbf{x}_{t}| y) =  \mu_\theta(\mathbf{x}_{t}|y) + s\cdot\Sigma_{\theta}(\mathbf{x}_t|y)\nabla logf_\Phi(y|\mathbf{x}_t, t)
\end{equation}

\begin{figure}[H]
    \centering
    \includegraphics[width=0.7\linewidth]{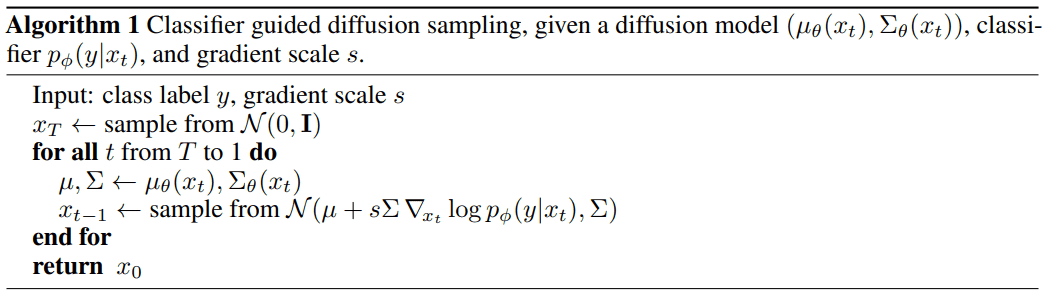}
    \caption{Algorithm of classifier guided diffusion. Source: ~\cite{NEURIPS2021_49ad23d1}}
    \label{fig:classifier-guidance-algo}
\end{figure}

\textbf{Classifier Free-Guidance}

Classifier-free guidance, proposed by Ho et al.~\cite{ho2022classifier}, allows for enhanced control in diffusion models by eliminating the need for separate classifiers. Instead of relying on a separate classifier, which increases training complexity and introduces bias potential, classifier-free guidance trains the diffusion model to directly learn and combine conditional and unconditional distributions during inference, streamlining the process.
In other words, the authors train a conditional diffusion model $\epsilon_{\theta}(\mathbf{x}_t|y)$ and an unconditional model $\epsilon_{\theta}(\mathbf{x}_t|y=0)$ as a single neural network as follows: 

\begin{equation}
    \hat{\epsilon}_{\theta}(x_t|y) = \epsilon_{\theta}(x_t|0) + s \cdot (\epsilon_{\theta}(x_t|y) - \epsilon_{\theta}(x_t|0))
\end{equation}

This approach is advantageous compared to the previous one as it trains a single model to guide the diffusion process and can take different types of conditional data such as text embeddings. We will see that many models rely on classifier free-guidance especially when training on multi-modal data. 

\subsubsection{Score-based generative models}
Score-Based Diffusion Models (SBDMs) are a class of diffusion models proposed by ~\cite{song2021score} that use score functions (gradient of the log probability density function) and Lagevin dynamics (iterative process where we draw samples from a distribution based only on its score function). Like GANs, SBDMs use adversarial training and try to generate images that are indistinguishable from real images. 

Instead of learning a probability density $p(\mathbf{x)}$, the neural network $s_\theta$ estimates the score function $\nabla_xlog p(\mathbf{x)}$ directly, and the training objective can be as follows: 

\begin{equation}
    \mathbb{E}_{p_\mathbf{x}}  \left[ \|\nabla_\mathbf{x} \log p(\mathbf{x}) - s_{\theta}(\mathbf{x}) -  \|^2_2 \right] = \int p(\mathbf{x}\| \nabla_\mathbf{x} \log p(\mathbf{x}) - s_{\theta}(\mathbf{x})\|^2_2 d\mathbf{x}
\end{equation}

While Langevin dynamics can sample $p(\mathbf{x})$ using the approximated score function, directly estimating $\nabla_xlogp(\mathbf{x}$ is difficult and imprecise. To address this, diffusion models learn score functions at various noise levels, achieved by perturbing the data with multiple scales of Gaussian noise.

So given the data distribution $p(x)$, we perturb it with Gaussian noise $N(0, \sigma_i^2 I)$ where $i = 1,2,\dots,L$ to obtain a noise-perturbed distribution:  
\begin{equation}
p_{\sigma_i}(\mathbf{x}) = \int p(\mathbf{y}) N(\mathbf{x}; \mathbf{y}, \sigma_i^2 I) \, d\mathbf{y}    
\end{equation}

Then we train a network $s_{\theta}(x, i)$, known as the Noise Conditional Score-Based Network (NCSN), to estimate the score function $\nabla_x \log p_{\sigma_i}(x)$ . The training objective is a weighted sum of Fisher divergences for all noise scales:  
\begin{equation}
    \sum_{i=1}^{L} \lambda(i) \mathbb{E}_{p_{\sigma_i}(x)}  
\left[ \| \nabla_x \log p_{\sigma_i}(x) - s_{\theta}(x, i) \|_2^2 \right]
\end{equation}

The authors of ~\cite{song2021score} combine components of NSCNs and DDPMs into one generative model, based on Stochastic Differential Equations (SDE) that does not depend on the data and no trainable parameters. Rather than perturbing data with a finite set of noise distributions, we utilize a continuous range of distributions that evolve over time through a diffusion process $\{\mathbf{x}_t\}_{t\in [0,T]}$. This process is governed by a predefined SDE. Then, it is possible to generate new samples by reversing this process. 
The forward process going from an input image $x_0$ to random noise $x_T$ is defined such that:
\begin{equation}
    d\mathbf{x} = \mathbf{f}(\mathbf{x}, t)dt + g(t)d\mathbf{w}
\end{equation}

Where:
\begin{itemize}
    \item $d\mathbf{w}$ is a Wiener process (random noise), 
    \item $\mathbf{f}(\mathbf{x},t)$ is the drift term and a vector valued function
    \item $\mathbf{g}(t)$ is the diffusion term and a scalar function
\end{itemize}

\begin{figure}[H]
    \centering
    \includegraphics[width=0.6\linewidth]{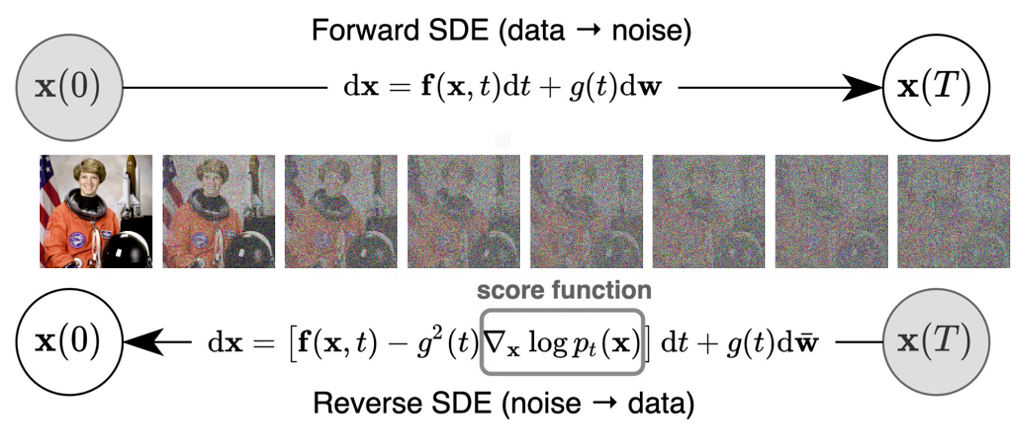}
    \caption{Score-based generative modeling through SDE}
    \label{fig:sde_modeling}
\end{figure}

After adding noise to the original data distribution for enough time steps, the perturbed distribution becomes close to a tractable noise distribution. Then it is possible to generate new samples by reversing the diffusion process and computing the reverse SDE given that the SDE was chosen to have a corresponding reverse SDE in closed form (see Figure ~\ref{fig:sde_modeling}):
\begin{equation}
    d\mathbf{x}  = \left[ \mathbf{f} (\mathbf{x} , t) - \mathbf{g}^2(t) \nabla_\mathbf{x}  \log p_t(\mathbf{x} ) \right] dt + \mathbf{g}(t) d\mathbf{w} 
\end{equation}

\subsection{Stable Diffusion and Latent diffusion models}

\begin{figure}[H]
    \centering
    \includegraphics[width=0.7\linewidth]{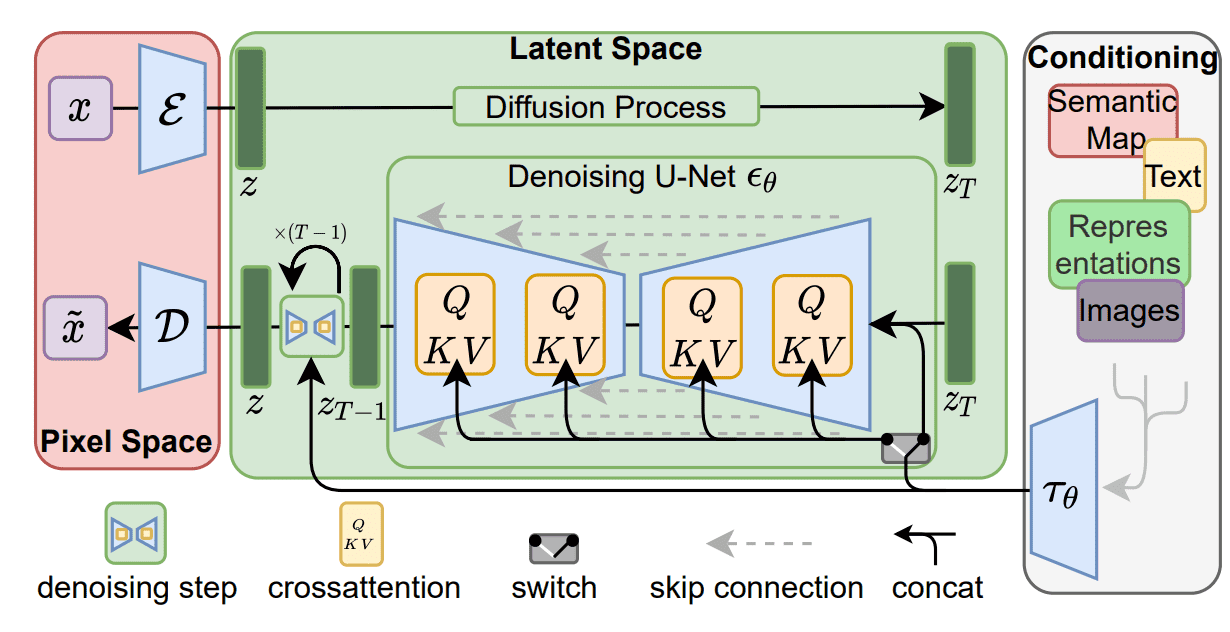}
    \caption{Latent diffusion architecture. Source: ~\cite{Kulkarni2023}}
    \label{fig:latent_diffusion}
\end{figure}

Latent diffusion models (LDMs) are yet another innovative extension of diffusion models ~\cite{9878449}. Instead of applying the diffusion on a high-dimensional input (pixel space), we project the input image into a smaller latent space and apply diffusion with latents as inputs. The authors of ~\cite{9878449} propose to use an encoder network $g$ to downsample the input into a latent representation $\mathbf{z}_t = g(\mathbf{x}_t)$ and apply the forward process to $\mathbf{z}_t $. Then the reverse process is the same as a standard diffusion process with a U-Net to generate new data that are then upsampled by a decoder network (see Figure ~\ref{fig:latent_diffusion}). 
Therefore, given an encoder $\varepsilon$ (Stable diffusion uses a pre-trained VAE encoder network), then the loss can be formulated as: 
\begin{equation}
    L_{LDM} = \mathbb{E}_{\varepsilon(x),t,\epsilon}[\| \mathbf{\epsilon} - \mathbf{\epsilon}_{\theta}(\mathbf{z}_t,t)^2\|]
\end{equation}

Stable diffusion can also be conditioned, in particular, using classifier-free guidance by adding conditional embeddings such as image features or text descriptions using a text encoder (e.g. CLIP's text encoder) to steer the generation process.

\subsubsection{Diffusion Transformers (DiT)}

One of the most recent diffusion-based models is the Diffusion Transformer (DiT) proposed in ~\cite{Peebles_2023_ICCV} which is an architecture that combines the principles of diffusion models and transformer models and that generates high-quality synthetic images.  It leverages the iterative denoising process inherent in diffusion models while utilizing the powerful representation learning capabilities of transformers for improved sample generation. The authors in ~\cite{Peebles_2023_ICCV} replace the U-Net backbone, in the LDM model, by a neural network called a Transformer ~\cite{NIPS2017_3f5ee243}. 
Transformers are a class of models based on self-attention mechanisms, and they have been proven to excel in tasks involving sequential data (like language processing). They work by attending to all input tokens at once and using multi-head self-attention to process the input efficiently. 
Mathematically, the attention mechanism can be formulated as follows:
\begin{equation}
    \textbf{Attention}(Q, K, V) = \text{softmax}\left( \frac{QK^T}{\sqrt{d_k}} \right) V
\end{equation}
Where:
\begin{itemize}
    \item \( Q \) (query), \( K \) (key), and \( V \) (value) are input representations.
    \item \( d_k \) is the dimensionality of the key vectors, and the softmax function normalizes the attention scores.
\end{itemize}

\begin{figure}[H]
    \centering
    \includegraphics[width=0.7\linewidth]{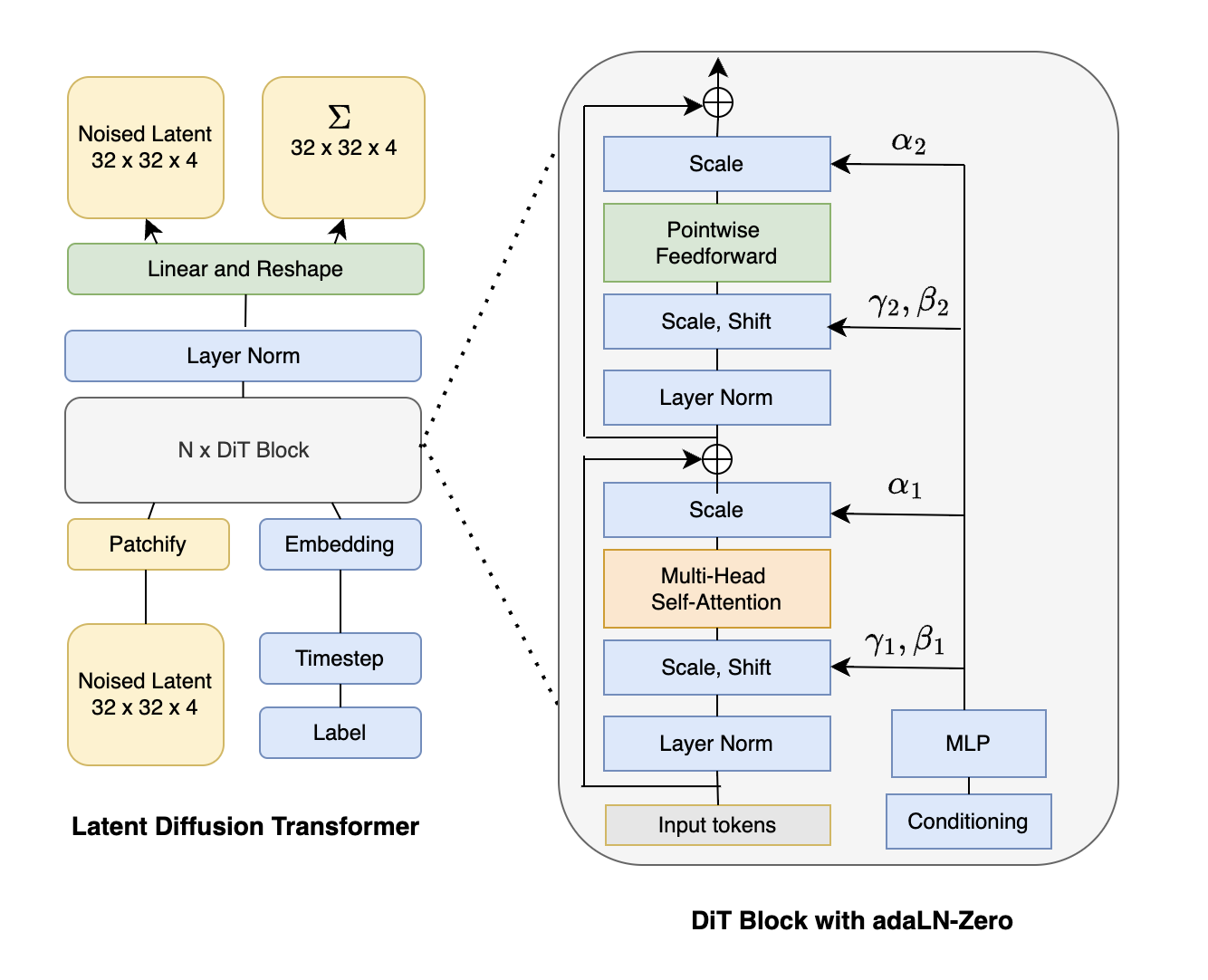}
    \caption{Architecture of the Diffusion Transformer}
    \label{fig:dit}
\end{figure}

In the context of a Diffusion Transformer (see Figure ~\ref{fig:dit}), the input to the transformer is typically a set of tokens or features (e.g., image patches, sequence tokens), and self-attention helps the model attend to dependencies across all tokens to capture long-range relationships. In the reverse process of the diffusion model, the transformer network is responsible for predicting the noise at each step, conditioned on the noisy data. For example, given the noisy image at time step \( t \), the transformer can model long-range spatial dependencies across the image patches (or sequence tokens) and generate a clean image at the next step:
\begin{equation}
    x_{t-1} = \textbf{Transformer}(\mu_\theta(x_t, t), \text{context})
\end{equation}

\vspace{-0.5cm}
Where:
\begin{itemize}
    \item \( \mu_\theta(x_t, t) \) is the predicted noise (as described in the reverse diffusion equation),
    \item \text{context} could be a conditioning input, such as a text prompt (in the case of text-to-image generation).
\end{itemize}

\section{Comparative Analysis}

We can define a set of metrics to evaluate a model family's general performance in image generation. Image Quality refers to the level of detail in the generated image. A model with high Image Quality strictly adheres to the imposed restrictions placed on it while maintaining a high level of detail, and absence of artifacts. A model with low Image Quality consistently generates images with large amounts of noise and/or artifacts and incoherent features\cite{Karras_2020_CVPR}. A model's Diversity refers to its range of potential outputs. A model with high diversity can produce a wide spectrum of images  while maintaining a constant image quality. A model with low diversity can only generate images in a narrow range with a constant image quality\cite{pmlr-v119-naeem20a}. Leaving this narrow range can lead to significant and rapid decreases in image quality. Controllability refers to how easy it is to guide the image generation process with some additional input. For example, if you wanted to generate variations of an image you could condition the model with an input image to help shape the generated image. A highly controllable model can take into account additional user input, understand the underlying features, and apply those features to the generated image. Training stability refers to the model's ability to reliably and smoothly converge over the training process.

Within the scope of generative models for image synthesis, Diffusion Models stand out for their ability to produce the highest quality images, often surpassing GANs, which also generate sharp visuals but may not achieve the same level of detail as diffusion-based approaches. VAEs, on the other hand, tend to yield blurrier images, indicating a trade-off in image fidelity. When it comes to diversity, both GANs and Diffusion Models excel at generating a wide variety of outputs, while VAEs can struggle with high variability, limiting their effectiveness in certain applications. In terms of controllability, Diffusion Models offer the most significant level of control over the generation process, allowing for precise adjustments, whereas GANs provide moderate to high control that can vary based on specific architectural choices. VAEs, however, exhibit limited controllability, making them less suitable for applications requiring fine-tuned image generation. Lastly, in terms of training stability, VAEs and Diffusion Models are generally more stable during the training process, reducing the likelihood of issues, while GANs often face challenges related to instability and mode collapse, which can hinder their performance and diversity\cite{10496521}. Table \ref{tab:comparison_models} summarizes aspects about image quality, diversity, controllability and training stability.

\begin{table}[h!]
    \centering
    \caption{Comparison of VAEs, GANs, and Diffusion Models for Text-to-Image Generation}
    \begin{tabular}{|p{2cm}|p{3cm}|p{3cm}|p{3cm}|p{3cm}|} 
        \hline
        \textbf{Model Type} & \textbf{Image Quality} & \textbf{Diversity} & \textbf{Controllability} & \textbf{Training Stability} \\
        \hline
        Variational Autoencoders (VAEs) & Moderate to High: Generally produces images with good quality but can be blurry due to the loss function used. & Moderate: Capable of generating diverse images but may struggle with high variability in complex datasets. & Moderate: Can condition on text embeddings but lacks fine-grained control over image features. & High: More stable during training compared to GANs, but can suffer from issues like posterior collapse. \\
        \hline
        Generative Adversarial Networks (GANs) & High: Known for generating sharp and detailed images, especially with techniques like Progressive Growing GANs. & High: Capable of producing a wide variety of images, especially with diverse training data. & Moderate to High: Can implement various conditioning methods (e.g., text-to-image) but may require complex architectures for precise control. & Moderate: Training can be unstable and sensitive to hyperparameters; mode collapse can occur, leading to reduced diversity. \\
        \hline
        Diffusion Models & Very High: Achieves state-of-the-art image quality, often surpassing GANs and VAEs in realism and detail. & High: Generates diverse images effectively, with the potential for high variability. & High: Allows for more explicit control over the generation process through iterative denoising steps and conditioning. & High: Generally more stable than GANs during training, with well-defined training objectives that reduce issues like mode collapse. \\
        \hline
    \end{tabular}
    \label{tab:comparison_models}
\end{table}

Scientific images can be detailed and high-resolution as many of them are acquired using advanced instruments, e.g., microscopes. In order to generate valuable synthetic images to augment scientific datasets, image quality is expected to be higher than in other domains, such as art. For example, MRI scans of human brains must both be detailed and expressly go through the HIPAA guidelines. Being able to generate synthetic MRI brains scans represent an invaluable opportunity to create a more diverse datasets from a few ``approved'' images, which could be used by researchers to train models\cite{10493074}\cite{10.1007/978-3-031-18576-2_12}\cite{moghadam2022morphologyfocuseddiffusionprobabilistic}. The challenge with scientific image generation lies in the Controllability or controlling their generation since pre-existent models are typically trained on data dissimilar to specialized imagery like microscopy data. 
So if you tried to just condition on a single cross-section on a standard GAN or Diffusion Model the results are likely lackluster. Alternatively, training a model from scratch would require a large dataset, which is actually the motivation for using image generation in the first place. Gathering sufficient amounts of data coming from experimental settings is often difficulty, and sometimes impossible, but without the sufficient quantity to reach convergence during training, the models can be useless. Considering the aforementioned strenghts, Diffusion Models are expected to exhibit optima performance in the synthesis of scientific imagery, as they address each of these criteria.

\section{Verification \& Validation}
Hallucinations and unexpected outcomes are some of the issues associated with GenAI. Other problems include inherent biases within training datasets can skew the generated images, reinforcing existing misconceptions~\cite{Lucas} or overlooking important, yet underrepresented, scientific phenomena. Validation becomes exceptionally difficult when dealing with completely novel scenarios, as there may be no existing experimental or observational data for comparison. This lack of ground truth poses risks of generating misleading visualizations that could inadvertently guide research down unproductive paths, therefore only rigorous scrutiny and expert validation could potentially mitigate these risks~\cite{Maleki:2024}

Verification and validation (V\&V) are essential for establishing the reliability and accuracy of AI generative models, and several efforts have focused on creating standardized benchmarks~\cite{rogerioIBM}, however curated datasets using scientific imaging are either extremely narrow~\cite{Rezende:nature:2021} or sparse~\cite{zhao2020}. Verification assesses a model's adherence to specified requirements and its performance under defined conditions. This includes unit testing for component correctness and performance evaluation against benchmark datasets. Cross-validation further examines the predictive performance across data subsets, indicating robustness. Validation determines whether the model accurately reflects real-world phenomena. In scientific imaging, validation involves qualitative expert (domain scientist) evaluations of generated image realism and quantitative metrics such as structural similarity index (SSIM) or peak signal-to-noise ratio (PSNR) for image quality. Incorporating domain-specific knowledge strengthens reliability. For example, in biological or material sciences imaging, comparisons against existing scientific models and datasets ensure generated outputs are both visually and scientifically sound. Through rigorous V\&V, researchers can avoid major pitfalls of generative AI models, and potentially model utilization in critical scientific applications.

\section{Conclusion \& Future Directions}

The future of text-to-image and image-to-image technologies promises significant advancements, with profound implications across diverse fields, notably scientific data analysis. We can anticipate continuous refinements in diffusion models, leading to hyper-realistic image generation coupled with increasingly granular control over specific attributes and detail. AI models will develop a deeper comprehension of contextual relationships, enabling the production of more nuanced and precise visual representations. Additionally, ongoing optimization of algorithms and hardware will yield faster generation times and reduced computational costs, while cloud-based platforms and mobile applications could democratize access to these technologies. A significant trend is the rapid progression of light-weight multimodal models~\cite{rogerioIBM,dai2024nvlm}, with expectations of substantial improvements in quality and coherence, particularly taking advantage of high-performance computer systems. Finally, AI will increasingly personalize image generation, learning individual user preferences to produce highly tailored visual outputs.

The impact of these technologies on scientific data analysis, particularly with scarce image sets from specialized instruments, will be transformative. AI-driven data augmentation promises to enable the generation of synthetic data to supplement limited datasets, enhancing the training of machine learning models for critical tasks like image segmentation and object detection. Moreover, AI will translate abstract scientific data into intuitive visual representations, facilitating the identification of patterns and trends in fields like genomics and materials science. By generating visual representations of potential scenarios, AI will assist scientists in formulating hypotheses and designing experiments, such as simulating molecular interactions or astronomical phenomena. AI can also be employed to identify and rectify errors in scientific images, improving the accuracy and reliability of data analysis. Furthermore, AI will foster increased collaboration by creating easily understandable visual representations of data for diverse scientific audiences.

Despite the immense potential, challenges remain. AI models can inherit biases from training data, leading to inaccurate results, which requires careful attention to dataset representativeness. The ``black box'' nature of some AI models poses challenges to interpretability, requiring efforts to develop more transparent models for scientific applications. Crucially, validation of AI-generated results against experimental data and established scientific principles is essential, especially when dealing with scarce datasets, to ensure the responsible and effective application of these powerful tools.

\section{Disclosure}
This article incorporates text and table generation facilitated by generative artificial intelligence. Although these tools aided in the organization and presentation of information, the authors retain full responsibility for the accuracy and scientific validity of the content. This article is a working in progress and further version will be shared shortly.

\section{Acknowledgments}
This work was supported by the US Department of Energy (DOE) Office of Science Advanced Scientific Computing Research (ASCR) and Basic Energy Sciences (BES) under Contract No. DE-AC02-05CH11231 to the Center for Advanced Mathematics for Energy Research Applications (CAMERA) program. It also included partial support from the DOE ASCR-funded project Autonomous Solutions for Computational Research with Immersive Browsing \& Exploration (ASCRIBE) and Laboratory Directed Research \& Development (LDRD) Program and the project Analytics through Diffusion Transformer Models (ADTM) for Scientific Image and Text.

\section*{Competing interests}
The authors declare that they have no competing interests.

\section*{Additional information}
Correspondence and material requests should be addressed to D.U.

\newpage
\bibliographystyle{plain} 
\bibliography{references.bib}

\begin{thebibliography}{10}

\bibitem{BetkerImprovingIG}
James Betker, Gabriel Goh, Li~Jing, † TimBrooks, Jianfeng Wang, Linjie Li,
  † LongOuyang, † JuntangZhuang, † JoyceLee, † YufeiGuo, †
  WesamManassra, † PrafullaDhariwal, † CaseyChu, † YunxinJiao, and Aditya
  Ramesh.
\newblock Improving image generation with better captions.
\newblock 2023.

\bibitem{dai2024nvlm}
Wenliang Dai, Nayeon Lee, Boxin Wang, Zhuolin Yang, Zihan Liu, Jon Barker,
  Tuomas Rintamaki, Mohammad Shoeybi, Bryan Catanzaro, and Wei Ping.
\newblock Nvlm: Open frontier-class multimodal llms.
\newblock {\em arXiv preprint arXiv:2401.03382}, 2024.

\bibitem{NEURIPS2021_49ad23d1}
Prafulla Dhariwal and Alexander Nichol.
\newblock Diffusion models beat gans on image synthesis.
\newblock In M.~Ranzato, A.~Beygelzimer, Y.~Dauphin, P.S. Liang, and J.~Wortman
  Vaughan, editors, {\em Advances in Neural Information Processing Systems},
  volume~34, pages 8780--8794. Curran Associates, Inc., 2021.

\bibitem{10493074}
Zolnamar Dorjsembe, Hsing-Kuo Pao, Sodtavilan Odonchimed, and Furen Xiao.
\newblock Conditional diffusion models for semantic 3d brain mri synthesis.
\newblock {\em IEEE Journal of Biomedical and Health Informatics},
  28(7):4084--4093, 2024.

\bibitem{foster2023generative}
David Foster.
\newblock {\em Generative Deep Learning: Teaching Machines to Paint, Write,
  Compose, and Play}.
\newblock O'Reilly Media, 2nd edition, 2023.

\bibitem{Gafni_Polyak_Ashual_Sheynin_Parikh_Taigman_2022}
Oran Gafni, Adam Polyak, Oron Ashual, Shelly Sheynin, Devi Parikh, and Yaniv
  Taigman.
\newblock Greater creative control for ai image generation, Jul 2022.

\bibitem{girdhar2024emuvideofactorizingtexttovideo}
Rohit Girdhar, Mannat Singh, Andrew Brown, Quentin Duval, Samaneh Azadi,
  Sai~Saketh Rambhatla, Akbar Shah, Xi~Yin, Devi Parikh, and Ishan Misra.
\newblock Emu video: Factorizing text-to-video generation by explicit image
  conditioning, 2024.

\bibitem{Goodfellow:2016}
Ian Goodfellow, Yoshua Bengio, and Aaron Courville.
\newblock {\em Deep Learning}.
\newblock MIT Press, 2016.

\bibitem{NIPS2014_5ca3e9b1}
Ian Goodfellow, Jean Pouget-Abadie, Mehdi Mirza, Bing Xu, David Warde-Farley,
  Sherjil Ozair, Aaron Courville, and Yoshua Bengio.
\newblock Generative adversarial nets.
\newblock In Z.~Ghahramani, M.~Welling, C.~Cortes, N.~Lawrence, and K.Q.
  Weinberger, editors, {\em Advances in Neural Information Processing Systems},
  volume~27. Curran Associates, Inc., 2014.

\bibitem{ho2022imagenvideohighdefinition}
Jonathan Ho, William Chan, Chitwan Saharia, Jay Whang, Ruiqi Gao, Alexey
  Gritsenko, Diederik~P. Kingma, Ben Poole, Mohammad Norouzi, David~J. Fleet,
  and Tim Salimans.
\newblock Imagen video: High definition video generation with diffusion models,
  2022.

\bibitem{NEURIPS2020_4c5bcfec}
Jonathan Ho, Ajay Jain, and Pieter Abbeel.
\newblock Denoising diffusion probabilistic models.
\newblock In H.~Larochelle, M.~Ranzato, R.~Hadsell, M.F. Balcan, and H.~Lin,
  editors, {\em Advances in Neural Information Processing Systems}, volume~33,
  pages 6840--6851. Curran Associates, Inc., 2020.

\bibitem{ho2022classifier}
Jonathan Ho and Tim Salimans.
\newblock Classifier-free diffusion guidance.
\newblock {\em arXiv preprint arXiv:2207.12598}, 2022.

\bibitem{rogerioIBM}
Leonid Karlinsky, Assaf Arbelle, Abraham Daniels, Ahmed Nassar, Amit Alfassi,
  Bo~Wu, Eli Schwartz, Dhiraj Joshi, Jovana Kondic, Nimrod Shabtay, Pengyuan
  Li, Roei Herzig, Shafiq Abedin, Shaked Perek, Sivan Harary, Udi Barzelay,
  Adi~Raz Goldfarb, Aude Oliva, Ben Wieles, Bishwaranjan Bhattacharjee, Brandon
  Huang, Christoph Auer, Dan Gutfreund, David Beymer, David Wood, Hilde Kuehne,
  Jacob Hansen, Joseph Shtok, Ken Wong, Luis~Angel Bathen, Mayank Mishra,
  Maksym Lysak, Michele Dolfi, Mikhail Yurochkin, Nikolaos Livathinos, Nimrod
  Harel, Ophir Azulai, Oshri Naparstek, Rafael Teixeira~de Lima, Rameswar
  Panda, Sivan Doveh, Shubham Gupta, Subhro Das, Syed Zawad, Yusik Kim, Zexue
  He, Alexander Brooks, Gabe Goodhart, Anita Govindjee, Derek Leist, Ibrahim
  Ibrahim, Aya Soffer, David Cox, Kate Soule, Luis Lastras, Nirmit Desai, Shila
  Ofek-koifman, Sriram Raghavan, Tanveer Syeda-Mahmood, Peter Staar, Tal Drory,
  and Rogerio Feris.
\newblock Granite vision: a lightweight, open-source multimodal model for
  enterprise intelligence.
\newblock Preprint, 2024.
\newblock Available at: https://arxiv.org/abs/2502.09927.

\bibitem{Karras_2020_CVPR}
Tero Karras, Samuli Laine, Miika Aittala, Janne Hellsten, Jaakko Lehtinen, and
  Timo Aila.
\newblock Analyzing and improving the image quality of stylegan.
\newblock In {\em Proceedings of the IEEE/CVF Conference on Computer Vision and
  Pattern Recognition (CVPR)}, June 2020.

\bibitem{kingma2022autoencodingvariationalbayes}
Diederik~P Kingma and Max Welling.
\newblock Auto-encoding variational bayes, 2022.

\bibitem{kirillov2023segany}
Alexander Kirillov, Eric Mintun, Nikhila Ravi, Hanzi Mao, Chloe Rolland, Laura
  Gustafson, Tete Xiao, Spencer Whitehead, Alexander~C. Berg, Wan-Yen Lo, Piotr
  Doll{\'a}r, and Ross Girshick.
\newblock Segment anything.
\newblock {\em arXiv:2304.02643}, 2023.

\bibitem{Kokhlikyan_Jayaraman_Bordes_Guo_Chaudhuri_2023}
Narine Kokhlikyan, Bargav Jayaraman, Florian Bordes, Chuan Guo, and Kamalika
  Chaudhuri.
\newblock Emu: Enhancing image generation models using photogenic needles in a
  haystack, Sep 2023.

\bibitem{Kulkarni2023}
Akshay Kulkarni, Adarsha Shivananda, Anoosh Kulkarni, and Dilip Gudivada.
\newblock {\em Diffusion Model and Generative AI for Images}, pages 155--177.
\newblock Apress, 2023.

\bibitem{Lucas}
Jason~S. Lucas, Barani~Maung Maung, Maryam Tabar, Keegan McBride, and Dongwon
  Lee.
\newblock The longtail impact of generative ai on disinformation: Harmonizing
  dichotomous perspectives.
\newblock {\em IEEE Intelligent Systems}, 39(5):12--19, 2024.

\bibitem{Maleki:2024}
Negar Maleki, Balaji Padmanabhan, and Kaushik Dutta.
\newblock Ai hallucinations: A misnomer worth clarifying.
\newblock In {\em 2024 IEEE Conference on Artificial Intelligence (CAI)}, pages
  133--138, 2024.

\bibitem{mirza2014conditionalgenerativeadversarialnets}
Mehdi Mirza and Simon Osindero.
\newblock Conditional generative adversarial nets, 2014.

\bibitem{moghadam2022morphologyfocuseddiffusionprobabilistic}
Puria~Azadi Moghadam, Sanne~Van Dalen, Karina~C. Martin, Jochen Lennerz,
  Stephen Yip, Hossein Farahani, and Ali Bashashati.
\newblock A morphology focused diffusion probabilistic model for synthesis of
  histopathology images, 2022.

\bibitem{pmlr-v119-naeem20a}
Muhammad~Ferjad Naeem, Seong~Joon Oh, Youngjung Uh, Yunjey Choi, and Jaejun
  Yoo.
\newblock Reliable fidelity and diversity metrics for generative models.
\newblock In Hal~Daumé III and Aarti Singh, editors, {\em Proceedings of the
  37th International Conference on Machine Learning}, volume 119 of {\em
  Proceedings of Machine Learning Research}, pages 7176--7185. PMLR, 13--18 Jul
  2020.

\bibitem{nichol2021improved}
Alex Nichol and Prafulla Dhariwal.
\newblock Improved denoising diffusion probabilistic models.
\newblock In {\em Proceedings of the 38th International Conference on Machine
  Learning (ICML)}, volume 139 of {\em Proceedings of Machine Learning
  Research}, pages 8162--8171. PMLR, 2021.

\bibitem{openai2021clip}
{OpenAI}.
\newblock Clip: Connecting text and images.
\newblock \url{https://openai.com/index/clip/}, 2021.
\newblock Accessed: 02/27/2025.

\bibitem{Peebles_2023_ICCV}
William Peebles and Saining Xie.
\newblock Scalable diffusion models with transformers.
\newblock In {\em Proceedings of the IEEE/CVF International Conference on
  Computer Vision (ICCV)}, pages 4195--4205, October 2023.

\bibitem{10.1007/978-3-031-18576-2_12}
Walter H.~L. Pinaya, Petru-Daniel Tudosiu, Jessica Dafflon, Pedro~F. Da~Costa,
  Virginia Fernandez, Parashkev Nachev, Sebastien Ourselin, and M.~Jorge
  Cardoso.
\newblock Brain imaging generation with latent diffusion models.
\newblock In Anirban Mukhopadhyay, Ilkay Oksuz, Sandy Engelhardt, Dajiang Zhu,
  and Yixuan Yuan, editors, {\em Deep Generative Models}, pages 117--126, Cham,
  2022. Springer Nature Switzerland.

\bibitem{radford2021learningtransferablevisualmodels}
Alec Radford, Jong~Wook Kim, Chris Hallacy, Aditya Ramesh, Gabriel Goh,
  Sandhini Agarwal, Girish Sastry, Amanda Askell, Pamela Mishkin, Jack Clark,
  Gretchen Krueger, and Ilya Sutskever.
\newblock Learning transferable visual models from natural language
  supervision, 2021.

\bibitem{radford2016unsupervisedrepresentationlearningdeep}
Alec Radford, Luke Metz, and Soumith Chintala.
\newblock Unsupervised representation learning with deep convolutional
  generative adversarial networks, 2016.

\bibitem{ramesh2021zeroshottexttoimagegeneration}
Aditya Ramesh, Mikhail Pavlov, Gabriel Goh, Scott Gray, Chelsea Voss, Alec
  Radford, Mark Chen, and Ilya Sutskever.
\newblock Zero-shot text-to-image generation, 2021.

\bibitem{Rezende:nature:2021}
Mariana~T. Rezende, Raniere Silva, Fagner de~O. Bernardo, Alessandra H.~G.
  Tobias, Paulo H.~C. Oliveira, Tales~M. Machado, Caio~S. Costa, Fatima N.~S.
  Medeiros, Daniela~M. Ushizima, Claudia~M. Carneiro, and Andrea G.~C. Bianchi.
\newblock Cric searchable image database as a public platform for conventional
  pap smear cytology data.
\newblock {\em Nature Scientific Data}, 8(1):151, 2021.

\bibitem{9878449}
Robin Rombach, Andreas Blattmann, Dominik Lorenz, Patrick Esser, and Bjorn
  Ommer.
\newblock { High-Resolution Image Synthesis with Latent Diffusion Models }.
\newblock In {\em 2022 IEEE/CVF Conference on Computer Vision and Pattern
  Recognition (CVPR)}, pages 10674--10685, Los Alamitos, CA, USA, June 2022.
  IEEE Computer Society.

\bibitem{Sheynin_2024_CVPR}
Shelly Sheynin, Adam Polyak, Uriel Singer, Yuval Kirstain, Amit Zohar, Oron
  Ashual, Devi Parikh, and Yaniv Taigman.
\newblock Emu edit: Precise image editing via recognition and generation tasks.
\newblock In {\em Proceedings of the IEEE/CVF Conference on Computer Vision and
  Pattern Recognition (CVPR)}, pages 8871--8879, June 2024.

\bibitem{pmlr-v37-sohl-dickstein15}
Jascha Sohl-Dickstein, Eric Weiss, Niru Maheswaranathan, and Surya Ganguli.
\newblock Deep unsupervised learning using nonequilibrium thermodynamics.
\newblock In Francis Bach and David Blei, editors, {\em Proceedings of the 32nd
  International Conference on Machine Learning}, volume~37 of {\em Proceedings
  of Machine Learning Research}, pages 2256--2265, Lille, France, 07--09 Jul
  2015. PMLR.

\bibitem{song2021score}
Yang Song and Stefano Ermon.
\newblock Score-based generative modeling through stochastic differential
  equations.
\newblock In {\em Proceedings of the International Conference on Learning
  Representations (ICLR)}, 2021.

\bibitem{Spataro_2023}
Jared Spataro.
\newblock Introducing microsoft 365 copilot – your copilot for work, Mar
  2023.

\bibitem{Sun:2024}
Yujie Sun, Dongfang Sheng, Zihan Zhou, and Yifei Wu.
\newblock Ai hallucination: towards a comprehensive classification of distorted
  information in artificial intelligence-generated content.
\newblock {\em Humanities and Social Sciences Communications}, 11(1):1278,
  2024.

\bibitem{lifearchitect}
Alan~D. Thompson.
\newblock Life.architect, 2024.

\bibitem{NIPS2017_3f5ee243}
Ashish Vaswani, Noam Shazeer, Niki Parmar, Jakob Uszkoreit, Llion Jones,
  Aidan~N Gomez, \L~ukasz Kaiser, and Illia Polosukhin.
\newblock Attention is all you need.
\newblock In I.~Guyon, U.~Von Luxburg, S.~Bengio, H.~Wallach, R.~Fergus,
  S.~Vishwanathan, and R.~Garnett, editors, {\em Advances in Neural Information
  Processing Systems}, volume~30. Curran Associates, Inc., 2017.

\bibitem{8578241}
Tao Xu, Pengchuan Zhang, Qiuyuan Huang, Han Zhang, Zhe Gan, Xiaolei Huang, and
  Xiaodong He.
\newblock Attngan: Fine-grained text to image generation with attentional
  generative adversarial networks.
\newblock In {\em 2018 IEEE/CVF Conference on Computer Vision and Pattern
  Recognition}, pages 1316--1324, 2018.

\bibitem{8237891}
Han Zhang, Tao Xu, Hongsheng Li, Shaoting Zhang, Xiaogang Wang, Xiaolei Huang,
  and Dimitris Metaxas.
\newblock Stackgan: Text to photo-realistic image synthesis with stacked
  generative adversarial networks.
\newblock In {\em 2017 IEEE International Conference on Computer Vision
  (ICCV)}, pages 5908--5916, 2017.

\bibitem{zhao2020}
Kai Zhao, Sheng Di, Xin Lian, Sihuan Li, Dingwen Tao, Julie Bessac, Zizhong
  Chen, and Franck Cappello.
\newblock Sdrbench: Scientific data reduction benchmark for lossy compressors.
\newblock In {\em 2020 IEEE International Conference on Big Data (Big Data)},
  pages 2716--2724, 2020.

\bibitem{Zhou:2024}
Lexin Zhou, Wout Schellaert, Fernando Mart{\'\i}nez-Plumed, Yael Moros-Daval,
  C{\`e}sar Ferri, and Jos{\'e} Hern{\'a}ndez-Orallo.
\newblock Larger and more instructable language models become less reliable.
\newblock {\em Nature}, 634(8032):61--68, 2024.

\bibitem{10496521}
Jingyuan Zhu, Huimin Ma, Jiansheng Chen, and Jian Yuan.
\newblock High-quality and diverse few-shot image generation via masked
  discrimination.
\newblock {\em IEEE Transactions on Image Processing}, 33:2950--2965, 2024.

\end{thebibliography}

\end{document}